\documentclass[10pt,twocolumn,letterpaper]{article}

\usepackage{wacv}
\usepackage{times}
\usepackage{epsfig}
\usepackage{graphicx}
\usepackage{amsmath}
\usepackage{amssymb}
\usepackage{bm}

\usepackage[]{authblk}

\usepackage{framed,multirow}
\usepackage{latexsym}
\usepackage{booktabs} 
\usepackage[]{algorithm2e}
\usepackage{epigraph}
\usepackage{graphicx}
\usepackage{subcaption}
\usepackage{cite}
\usepackage{soul}
\usepackage{comment}

\usepackage{url}

\graphicspath{{fig/}}
\usepackage[pagebackref=true,breaklinks=true,letterpaper=true,colorlinks,bookmarks=false]{hyperref}

\wacvfinalcopy 

\def\wacvPaperID{600} 

\ifwacvfinal\fi
\begin{document}

\title{Recommending Outfits from Personal Closet}



\author[1]{Pongsate Tangseng}
\author[2]{Kota Yamaguchi}
\author[1,3]{Takayuki Okatani}

\affil[1]{Tohoku University}
\affil[2]{CyberAgent,Inc.}
\affil[3]{RIKEN Center for AIP}

\affil[1]{\tt\small {\{tangseng,okatani\}@vision.is.tohoku.ac.jp}}
\affil[2]{\tt\small {yamaguchi\_kota@cyberagent.co.jp}}

\maketitle
\ifwacvfinal\thispagestyle{empty}\fi

\begin{abstract}
 We consider grading a fashion outfit for recommendation, where we assume that users have a closet of items and we aim at producing a score for an arbitrary combination of items in the closet. The challenge in outfit grading is that the input to the system is a bag of item pictures that are unordered and vary in size. We build a deep neural network-based system that can take variable-length items and predict a score. We collect a large number of outfits from a popular fashion sharing website, \emph{Polyvore}, and evaluate the performance of our grading system. We compare our model with a random-choice baseline, both on the traditional classification evaluation and on people's judgment using a crowdsourcing platform. With over 84\% in classification accuracy and 91\% matching ratio to human annotators, our model can reliably grade the quality of an outfit. We also build an outfit recommender on top of our grader to demonstrate the practical application of our model for a personal closet assistant.
\end{abstract}

\section{Introduction}





There have been growing interests in applying computer vision to fashion, perhaps due to the rapid advancement in computer vision research~\cite{fashionCVPR12, paperdoll, HipsterWarsECCV14, CFPD, CRFClothesParsing, CCP, runway_to_realway,Veit2015, Oramas2016,Hsiao2017}. 
One of the popular fashion applications is item recommendation~\cite{magic_closet, collaborative_fashion_recommendation, street_to_shop, mining_outfit}, where the objective is to suggest items to users based on user's and/or society's preference. Computer vision is used in various fashion applications such as e-commerce and social media. 
Recently, Amazon announced their automatic style assistant called ``Echo Look\textsuperscript{\texttrademark}''. Although the underlying mechanism is not published, emerging commercial applications confirm the ripe of computer vision applications in fashion.

   
\begin{figure}[t]
\centering
  \includegraphics[width=0.6\columnwidth]{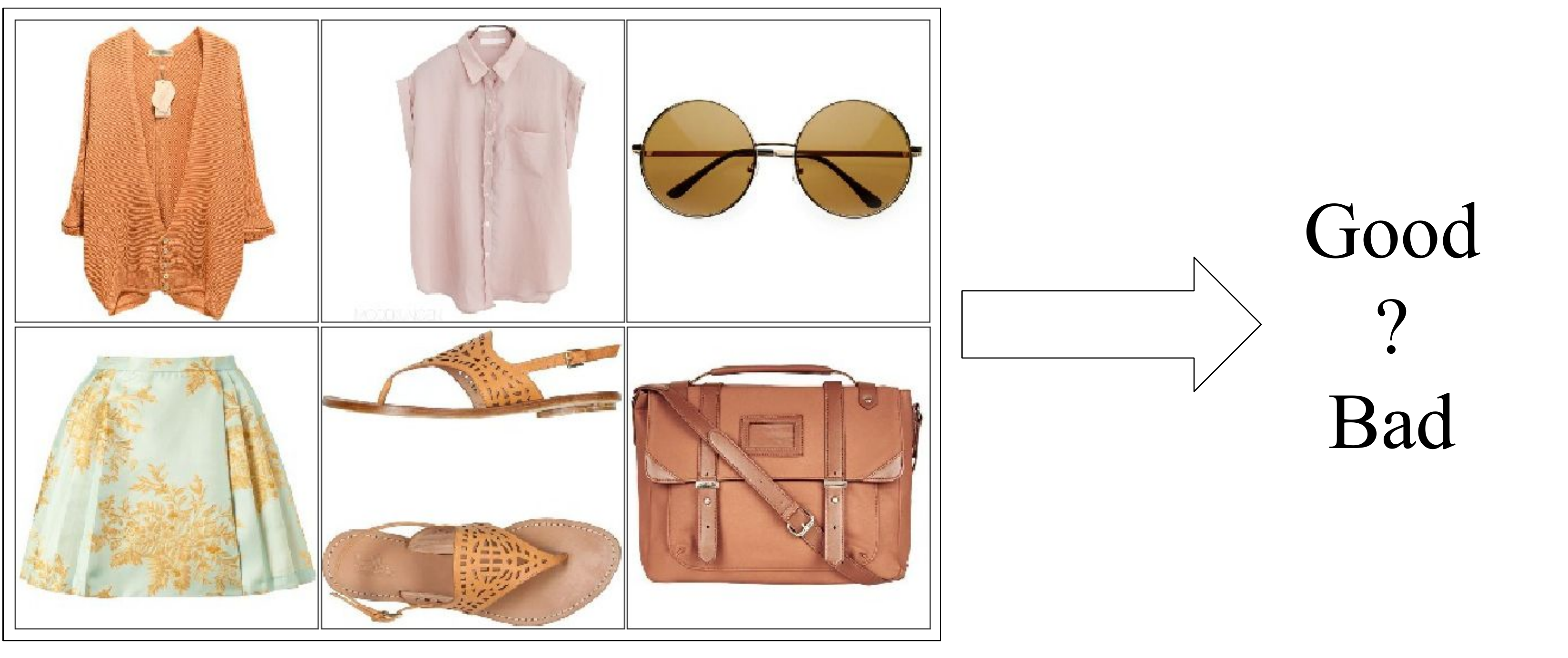}
  \vspace{-5pt}
  \caption{Given an arbitrary number of items, our goal is to evaluate the quality of the outfit combination.}
  \label{fig:outfit_problem}
  \vspace{-10pt}
\end{figure}

Measuring the quality of outfit is essential in building fashion recommendation system. In this paper, we consider the problem of grading arbitrary combination of items as a whole (Figure~\ref{fig:outfit_problem}). 
Previous works in outfit evaluation can be divided into two groups based on the input format: a worn outfit as a full-body picture as in~\cite{dressup, fashion144k, 128floats, HipsterWarsECCV14}, and as a set of images of items~\cite{collaborative_fashion_recommendation, mining_outfit}, or a combination of both~\cite{magic_closet}. Each outfit can have an arbitrary number of items. For examples, in one day, one might prefer a combination of a jacket, a t-shirt, and jeans, while in the another she might want to wear a dress. Our goal is to build a machine learning system that accepts a variable numbers of items yet produce a consistent score for any size of combinations.

In this paper, we view an outfit as a bag of fashion items and utilize deep neural networks to produce a score for a fixed-length representation of outfits. 
Unlike style recognition~\cite{dressup, HipsterWarsECCV14, 128floats, Hsiao2017}, we take item images in isolation, not on human body, as seen on e-commerce sites or catalogs. We collect a large number of outfit data from a popular fashion website \url{polyvore.com}, and evaluate our approach based on standard classification metrics and human judgment. Our Polyvore409k dataset consists of 409,776 sets of clothing items from 644,192 unique items. The dataset forms a large bipartite graph of items and outfits. We partition the dataset into training and testing sets such that there is no overlapping nodes and edges between the sets, and use them measure the classification performance. We also conduct a human study using crowdsourcing to assess predicted scores against human judgment, and show our model closely resembles human behavior. Using our grader, we build an outfit recommendation system that takes clothing items as an input and 
recommends the best outfits from the given items, to demonstrate the usefulness in a real-world scenario of personal outfit assistant. The contributions of the paper are summarized below:
\begin{enumerate}
  \vspace{-5pt}
\itemsep-0.3em
\item We build Polyvore409k dataset containing 409,776 outfits and 644,192 items. Every outfit covers the entire body with a variable numbers of items.
\item We propose an outfit grader that produces a score for fashion outfits with a variable number of items. Our empirical study shows that our model achieves 84\% of accuracy and precision in Polyvore409k dataset.
\item We propose a human judgment framework on outfit quality, which provides a simple and reliable method to verify the reliability of outfit verifiers using a crowdsourcing platform.
\item We demonstrate that our outfit grader can build a recommendation system that suggests good outfits from a pool of items.
  \vspace{-5pt}
\end{enumerate}



\section{Related Work} \label{sec:related_works}



\subsection{Outfit Modeling}
The use of computer vision techniques to study fashion is gaining popularity. Some early studies work on outfit images~\cite{dressup, fashion144k, 128floats}. Although these studies can use the appearances of outfit on a real subject, accurately identifying items in an outfit image is still an open problem. The manual annotation is costly, and the automatic detection and segmentation of fashion items in an outfit image \cite{fashionCVPR12, paperdoll, CCP, CRFClothesParsing, CFPD, tangseng} are still not reliable. For example, ``dress'' and ``top with skirt'' are often incorrectly segmented, and the small objects like shoes and accessories are often missed. In addition, the importance of each item to the overall style may or may not be related to the scale of the item in an outfit image. In this paper, we aim to study outfits as a combination of items, where each item has its own image.

Some studies treat outfits as combinations of item images as well~\cite{magic_closet, amazon_data, collaborative_fashion_recommendation, mining_outfit}. In \cite{magic_closet}, items in an outfit are recommended according to the requested occasion and the existing items in that outfit. The work by \cite{collaborative_fashion_recommendation} focuses on learning personal preference on fashion based on accounts and associated outfits from \url{polyvore.com}. A study of pairwise relationship between fashion items was explored in \cite{amazon_data} using co-purchase data. The work by \cite{mining_outfit} also uses data from \url{polyvore.com} to learn outfits as combinations of items based on item image, name, and category. They create an item recommendation system that suggests an item to match with other manually selected items.

Outfits have a natural structure based on human body, but \cite{magic_closet, collaborative_fashion_recommendation, mining_outfit} consider outfits with fixed number of items without considering variation in the structure. Outfits in \cite{collaborative_fashion_recommendation} consist of one top, one bottom, and a pair of shoes without considering full-body items such as a dress, nor accessories. In \cite{magic_closet}, recommendation is made for either whole outfit, or upper-lower body pairs. Likewise, outfits in \cite{mining_outfit} consist of 4 items, regardless of item role. \cite{mining_outfit} does not guarantee the completeness of outfits. Since an outfit can be a collection of any items, it is possible to have incomplete outfits that do not cover whole body, such as an outfit consisting only of four pairs of boots.

In this work, we view outfits as collections of items from \url{polyvore.com}, similar to \cite{collaborative_fashion_recommendation, mining_outfit}. We arrange the outfit data such that each outfit covers the entire body by considering the body part covered by each item, with variable number of items in the outfit.

\begin{table*}
\small
\centering
\caption{Comparison of outfit datasets}
\label{table:approaches_comparison}
\begin{tabular}{|p{4cm}|p{2.2cm}|p{3.7cm}|p{2cm}|p{3.7cm}}
\hline
& \cite{magic_closet} & \cite{collaborative_fashion_recommendation} & \cite{mining_outfit} & \multicolumn{1}{p{3cm}|}{\textbf{Ours}} \\ \hline
\textbf{\#images in dataset} & 24,417 & 85,252 & 347,339 & \multicolumn{1}{l|}{644,192} \\ \hline
\textbf{Annotation method} & Crowdsourcing & \multicolumn{3}{l|}{Metadata from polyvore.com} \\ \hline
\textbf{Outfit labels} & Occasions and attributes & \parbox[t]{3.7cm}{User-created: positive\\Randomly created: negative} & Based on votes & \multicolumn{1}{l|}{\parbox[t]{3.7cm}{User-created: positive\\Randomly created: negative}} \\ \hline
\textbf{\#items in outfits} & 2 & 3 & 4 & \multicolumn{1}{l|}{Variable, up to 8} \\ \hline
\textbf{Human body parts} & 2 parts & 3 parts & No & \multicolumn{1}{l|}{6 parts} \\ \hline
\textbf{Train/test item separation} & \multicolumn{2}{l|}{Not verified} & \multicolumn{2}{l|}{Verified} \\ \hline
\textbf{Evaluated by human} & No & No & No & \multicolumn{1}{l|}{Yes} \\ \hline
\end{tabular}
\end{table*}

\subsection{Fashion Datasets}\label{sec:existing_dataset}

The number of fashion datasets is growing. Each image in some datasets~\cite{paperdoll, CFPD, CCP, fashion144k} is an outfit images, which contains many items. In other datasets~\cite{street_to_shop, amazon_data,collaborative_fashion_recommendation, mining_outfit}, each image contains only one item. Some datasets~\cite{magic_closet, deepfashion} are combinations of both type.
In \cite{amazon_data}, item combinations come in pairwise format from amazon co-purchase data. However, items that are bought together do not necessarily mean that they look good together as an outfit. 

There are segmentation datasets \cite{fashionCVPR12, CCP, CFPD} that seem suitable for our problem setup, because the datasets provide outfit images with the boundary of each item, but the number of samples is too few to learn a reasonable predictive model. Although \cite{collaborative_fashion_recommendation} and \cite{mining_outfit} use datasets with combination of images as outfits and each item has its own image, the dataset is not publicly available. For the above reasons, we collect and build a new dataset, Polyvore409k dataset, which we describe in section \ref{sec:dataset}.

In fashion outfit problem, each sample is an outfit which is a combination of items. We have to cleanly separate training data from testing data both for a set and individual items. \cite{magic_closet} and \cite{collaborative_fashion_recommendation} do not describe the detail on separation. \cite{mining_outfit} constructs a graph dataset, where each node represents an outfit, and a connection between any two nodes is formed if these two outfits have a common items. After that, the graph is segmented based on connected components. In this work, we use an efficient alternative approach to split a graph, which we describe in section \ref{sec:train_test_split}.

\section{Polyvore409k Dataset} \label{sec:dataset}
This section describes our Polyvore409k dataset that consists of variable-length sets of items. Our Polyvore409k dataset has 409k outfits consisting of 644k item images. The comparison of outfit datasets to the previous work is shown in table \ref{table:approaches_comparison}. 
We plan to release the metadata of items and outfits, including the URLs to the images, to the public.



\subsection{Data Collection}\label{sec:data_collection}



We collect Polyvore409k dataset from the fashion-based social media website \url{polyvore.com}. Each outfit, or \emph{set} in Polyvore's terminology, consists of a title, items in the set, a composed image, and behavioral data such as likes and comments from other users. 

\subsection{Data Preprocessing}\label{sec:data_preprocessing}

\paragraph{Data Cleansing}
The collected sets can contain non-clothing items or items that cannot be worn such as logo, background image for presentation purpose, or cosmetic items.
We remove the item if its name does not contain clothing categories. After that, each item in a set is categorized into one of 6 outfit parts according to its categories:
\begin{enumerate}
  \vspace{-5pt}
  \itemsep-0.3em
  \item Outer: coat, jacket, parka, etc.
  \item Upper-body: blouse, shirt, polo, etc.
  \item Lower-body: pants, jeans, skirt, joggers, etc.
  \item Full-body: dress, gown, jumpsuit, robe, etc.
  \item Feet: shoes, boots, flats, clutches, etc.
  \item Accessory: bag, glove, necklace, earring, etc.
  \vspace{-5pt}
\end{enumerate}

Our definition of an outfit is a set that covers both upper and lower part of body, each of first 5 categories has at most one item, and at most three items for accessory. Sets that do not cover the whole body, e.g. missing lower body, are removed. At the end, we obtained 409,776 valid outfits which are composed of 644,192 unique items.

We consider only two layers on the upper body (outer and upper) because of the visibility, as the layers under two outermost layers are usually covered. We process sweaters, knitwear, and the likes as outer-upper hybrids. They will be considered as an upper if the outfit has other outer, and as an outer if the outfit does not have. The list of item categories and respective outfit parts is included as supplementary.


\renewcommand{\tabcolsep}{3pt}
\begin{table}
  \centering
  \small
  \caption{Number of unique items in each outfit part}
  \label{table:nitem_each_part}
\begin{tabular}{@{}lcccccc@{}}
\toprule
Part & Outer & Upper & Lower & Full & Feet & Accessory \\ \midrule
Train & 11,168 & 21,760 & 16,287 & 11,523 & 26,574 & 60,760 \\
Test & 6,656 & 12,744 & 11,089 & 8,871 & 17,564 & 37,988 \\ \bottomrule
\end{tabular}
\end{table}

\subsection{Quality Measurement}
Measuring the quality of an outfit is a challenging task due to the subjective nature of judging visual appearance. The approach of \cite{mining_outfit} directly uses the number of votes (or \emph{like} in \url{polyvore.com}'s terminology) of the outfit on the website as a quality measurement. However, some studies \cite{chic_or_social, fashion144k, 128floats} argue that the number of votes from social media does not directly reflect the quality of the outfit, because the number of clicks is affected by a variety of factors, such as the topology of the social networks or the time when the outfit was published. In \cite{collaborative_fashion_recommendation}, the quality is defined by the preference of each user: outfits created by the user are treated as positive samples, and outfits created by randomly pick items are treated as negative. Given these insights, we take the following strategy.

\paragraph{Positive Samples}
Each Polyvore409k outfit has an associated \emph{like}s that Polyvore users provide.  Although the number of \emph{like} might not directly reflect the quality of the outfit, it still shows that some people like the outfit. As a result, we use 212,623 outfits that has least one \emph{like} as positive samples. In the future, as we obtain more data, we wish to increase the number of \emph{like} threshold.

\RestyleAlgo{boxruled}
\begin{algorithm}[t]
\footnotesize
\SetKwInOut{Input}{input}\SetKwInOut{Output}{output}
\Input{All outfits $\mathbf{O}$}
\Output{Set $A$, $B$, and $C$ containing outfits such that items in outfits in $A$ is not in $B$ and vice versa}
$A \leftarrow B \leftarrow C \leftarrow \emptyset$\;
$A \cup \{\mathbf{O}_0\}$\;
\For{$i\leftarrow 1$ \KwTo $|\mathbf{O}|$}{
   $O \leftarrow \mathbf{O}_i$\;
   $items_A \leftarrow$ items in outfits in $A$\;
    $items_B \leftarrow$ items in outfits in $B$\;
    $items_O \leftarrow$ items in $O$\;
    $sec_{AO} \leftarrow intersection(items_A,items_O)$\;
    $sec_{BO} \leftarrow intersection(items_B,items_O)$\;
    \lIf{$|sec_{AO}| > 0$ and $|sec_{BO}| > 0$}{
      $C \cup \{O\}$
    }
    \lElseIf{$|sec_{AO}| > 0$}{
      $A \cup \{O\}$
    }
    \lElseIf{$|sec_{BO}| > 0$}{
      $B \cup \{O\}$
    }
    \Else{
      \lIf{$|A|/2 > |B|$}{
         $B \cup \{O\}$
        }\lElse{
         $A \cup \{O\}$
        }
    }
}
\caption{Disjoint Set Sampling}
\label{algo:train_test_split}
\end{algorithm}
\vspace{-5pt}

\paragraph{Negative Samples} \label{sec:negative_samples}
Similar to \cite{collaborative_fashion_recommendation}, we use outfits created by picking items randomly as negative samples. We believe that there are some preferred combinations of colors, textures, or shape of items in an outfit, and we assume that a randomly created outfit has very small chance to match those preferences.

For each positive sample, we create two identical samples as negative samples, because the number of preferred combinations is expected to be significantly lower than random combinations. Then, we replace items in those two negative samples with random items of the same parts from the same train/test item pool. Although this sampling strategy is not \emph{i.i.d.}, this approach guarantees the disjoint set property between training and testing sets, and tends to produce \emph{hard} negative examples that shares some items with positive counterpart. Also, the distribution of number of items and existences of outfit parts in samples are preserved.

\begin{figure}
\centering
  \includegraphics[width=.45\textwidth]{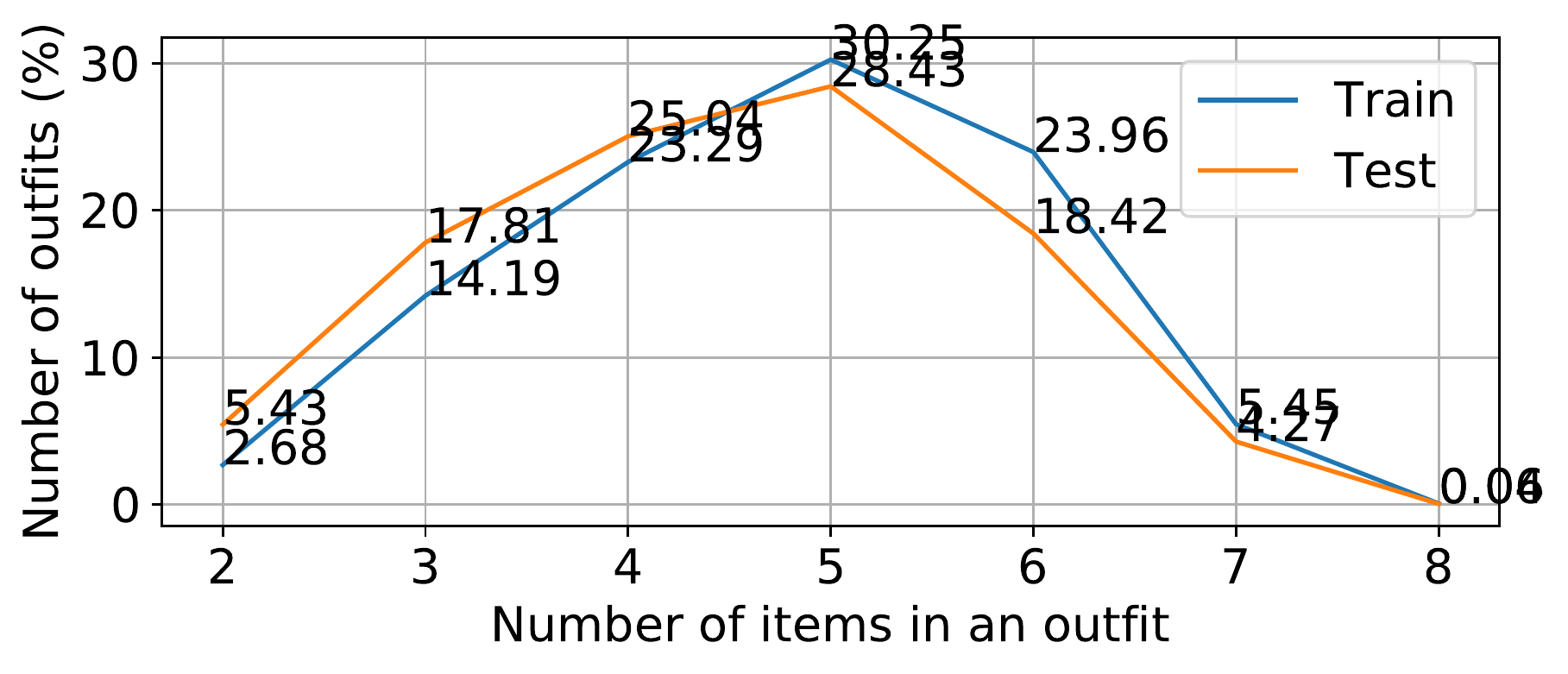}
  \vspace{-5pt}
  \caption{Distribution of number of items in outfits in training and testing partition are similar.}
  \label{fig:num_set_by_nitem}
\end{figure}

Table \ref{table:nitem_each_part} shows the number of items in each part of outfit. Figure~\ref{fig:num_set_by_nitem} shows the distribution of number of items in an outfit in train and test splits for both positive and negative samples.  Table \ref{table:nsamples_each_set} shows the numbers of positive and negative samples in each split.


\subsection{Evaluation Data}\label{sec:train_test_split}


The set-item relationship constitutes a bipartite graph, where nodes are outfits or items, and edges represent inclusion relationship.
For performance evaluation using Polyvore409k, we have to split the bipartite graph such that the training and testing splits do not share any item or outfits. We use Algorithm \ref{algo:train_test_split} to separate training and testing splits.

\begin{table}[t]
\centering
\small
\caption{Number of outfits in each train and test partition}
\label{table:nsamples_each_set}
\begin{tabular}{@{}lcc@{}}
\toprule
Number of outfits & Train & Test \\ \midrule
Positive samples & 66,434 & 26,813 \\
Negative samples & 132,868 & 53,626 \\ \midrule
Total & 199,302 & 80,439 \\ \midrule
Ratio positive:negative & 1:2 & 1:2 \\ \bottomrule
\end{tabular}

\end{table}

\section{Outfit Grader} \label{sec:outfit_verifier}

\begin{figure}[t]
\centering
  \includegraphics[width=0.8\columnwidth]{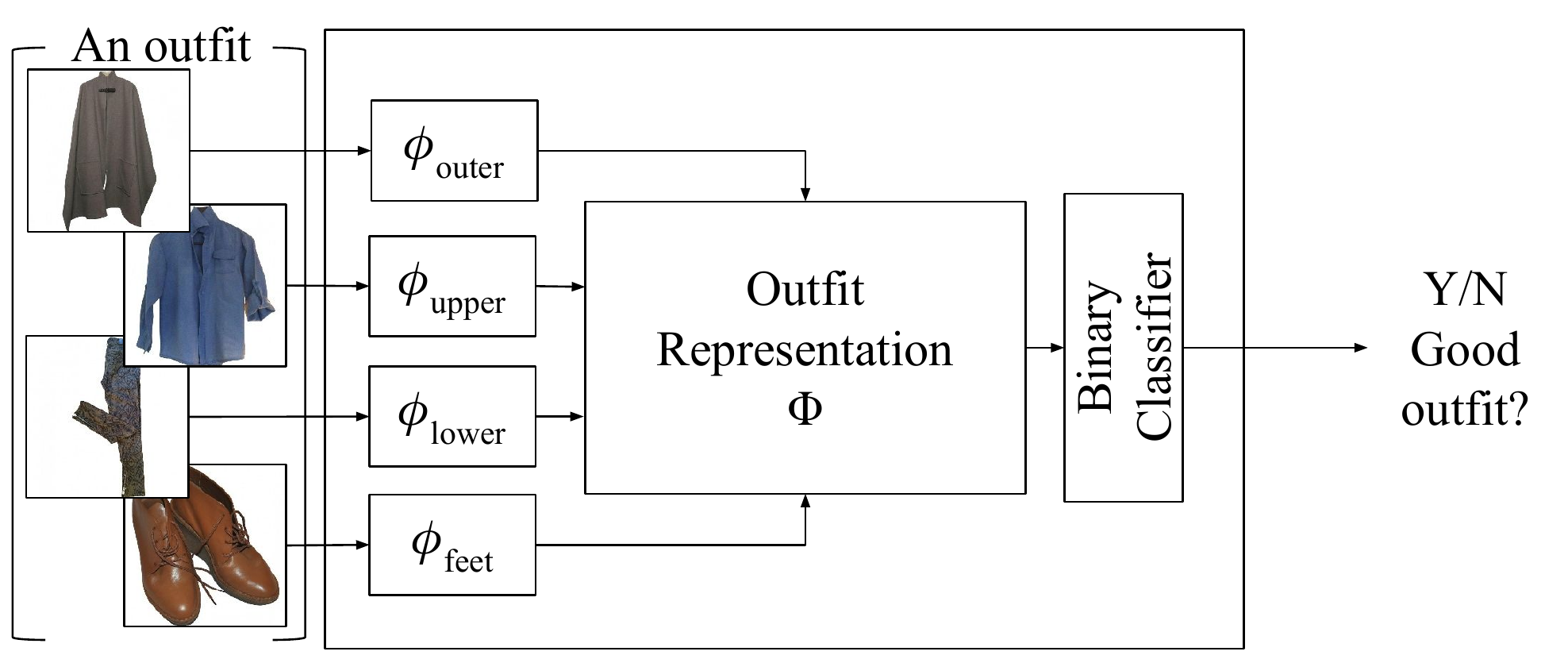}
  \vspace{-5pt}
  \caption{Outfit Grader}
  \label{fig:grader}
  \vspace{-5pt}
\end{figure}


\subsection{Problem Formulation}
We formulate the outfit grading as a binary classification problem. Given an outfit $O \equiv \{ x_\mathrm{outer}, x_\mathrm{upper}, \cdots, x_\mathrm{accessory3} \}$, where $x_\mathrm{part}$ is an item image, the goal is to learn a mapping function: $F: O \mapsto y$ to predict the outfit's quality $y \in \{0, 1\}$. Once we learn the mapping $F$, we are able to sort arbitrary combinations of items according to the prediction score.

The challenge is how to represent an outfit $O$ with a variable number of items. Luckily, the number of visible items is limited even though an outfit can contain a variable number of items. Therefore, we assign each item into one of the six categories and concatenate the item representations to produce the outfit representation. Figure~\ref{fig:grader} shows our grader. Our grader takes a bag of images and convert them to feature representations, then concatenates the individual features according to the item's category to produce the fixed-length representation. We describe details below.





\subsection{Item Representation}
We convert the image of each item in the outfit to a feature representation $\phi_\textrm{part}(x_\textrm{part})$, using a convolutional network. In this paper, we use ImageNet-pretrained ResNet-50~\cite{resnet}, and extract the 2,048-dimensional embedding from \texttt{pool5} layer as an item representation. We extract features for 5 item parts and up to 3 accessories. For missing parts, we give a mean image to obtain features which is equal to zero-input to the convolutional network.

\subsection{Outfit Representation}
After we extract features from each item, we concatenate all features in the fixed order to form an outfit representation $\Phi(O) \equiv [\phi_\textrm{outer}, \phi_\textrm{upper}, \cdots, \phi_\textrm{accessory2} ]$. Note that we allow accessories to appear multiple times in the outfit, and we simply concatenate all the accessory features ignoring the order. Outfits with less than 3 accessories get mean images as well to the other part. We have 5 item parts and 3 accessories per outfit, resulting in a 16,384 dimensional representation as the outfit representation.



\subsection{Scoring Outfits}
From the outfit representation $\Phi$, we learn a binary classifier and predict a score. We utilize a multi-layer perceptron (MLP) to learn the mapping function. 
In this paper, we compare 3 MLPs with various configurations to see the effect of number and size of fully-connected (FC) layers on this problem. The models we used are:


\begin{enumerate}
\vspace{-7pt}
\itemsep-0.3em
  \item one\_fc4096: one 4096-d FC layer 
  \item one\_fc128: one 128-d FC layer 
  \item two\_fc128: two 128-d FC layers 
  \vspace{-7pt}
\end{enumerate}

Each of fully-connected layers are followed by batch normalization and rectified linear activation (ReLU) with dropout. One 2-d linear layer followed by soft-max activation is added to every models to predict a score. We use multinomial logistic loss to learn the parameters of the grading model.





\section{Performance Evaluation} \label{sec:experiments}
\subsection{Evaluation Setup} \label{sec:item_image_evaluation}
We learn the grading model from the training split of Polyvore409k dataset, and evaluate the binary classification measures on the testing split. The performance is measured against the ground truth. In this paper, we report the performance of our models without fine-tuning the parameters of the convolutional network for the item feature extraction. We implement the neural network using Caffe framework~\cite{caffe}. We choose cross entropy as a loss function. We train the models for 400,000 iterations using stochastic gradient descent with momentum, where the initial learning rate and momentum are set to $10^{-4}$ and $0.9$, respectively.
We measure accuracy, precision, and recall to evaluate the performance. The prediction is counted as correct if it matches the ground truth.


\begin{table}[t]
\centering
\small
\caption{Accuracy, average precisions, and average recall of our outfit graders at 400,000 iterations.}
\label{table:accuracy_precision}
\begin{tabular}{@{}lccc@{}}
\toprule
 & Accuracy & Avg. Precision & Avg. Recall \\ \midrule
one\_fc4096 & \textbf{84.51} & \textbf{83.66} & \textbf{80.62} \\
one\_fc128 & 80.14 & 81.25 & 72.79 \\
two\_fc128 & 82.11 & 82.14 & 76.36 \\ \bottomrule
\end{tabular}
\end{table}

\begin{table}[t]
\centering
\small
\caption{Precision, recall, and F1 value of both classes from one\_fc4096 model at 400,000 iterations.}
\label{table:detail_metric}
\begin{tabular}{@{}l|ccc|c@{}}
\toprule
 & \multicolumn{3}{c|}{Testing} & Training \\ \midrule
 & Negative & Positive & Average & Average \\ \midrule
Precision & 85.60 & 81.73 & 83.66 & 99.25 \\
Recall & 92.29 & 68.95 & 80.62 & 99.31 \\
F1 & 88.82 & 74.80 & 81.81 & 99.28 \\ \bottomrule
\end{tabular}
\end{table}

\begin{figure}[t]
  \includegraphics[width=0.45\textwidth]{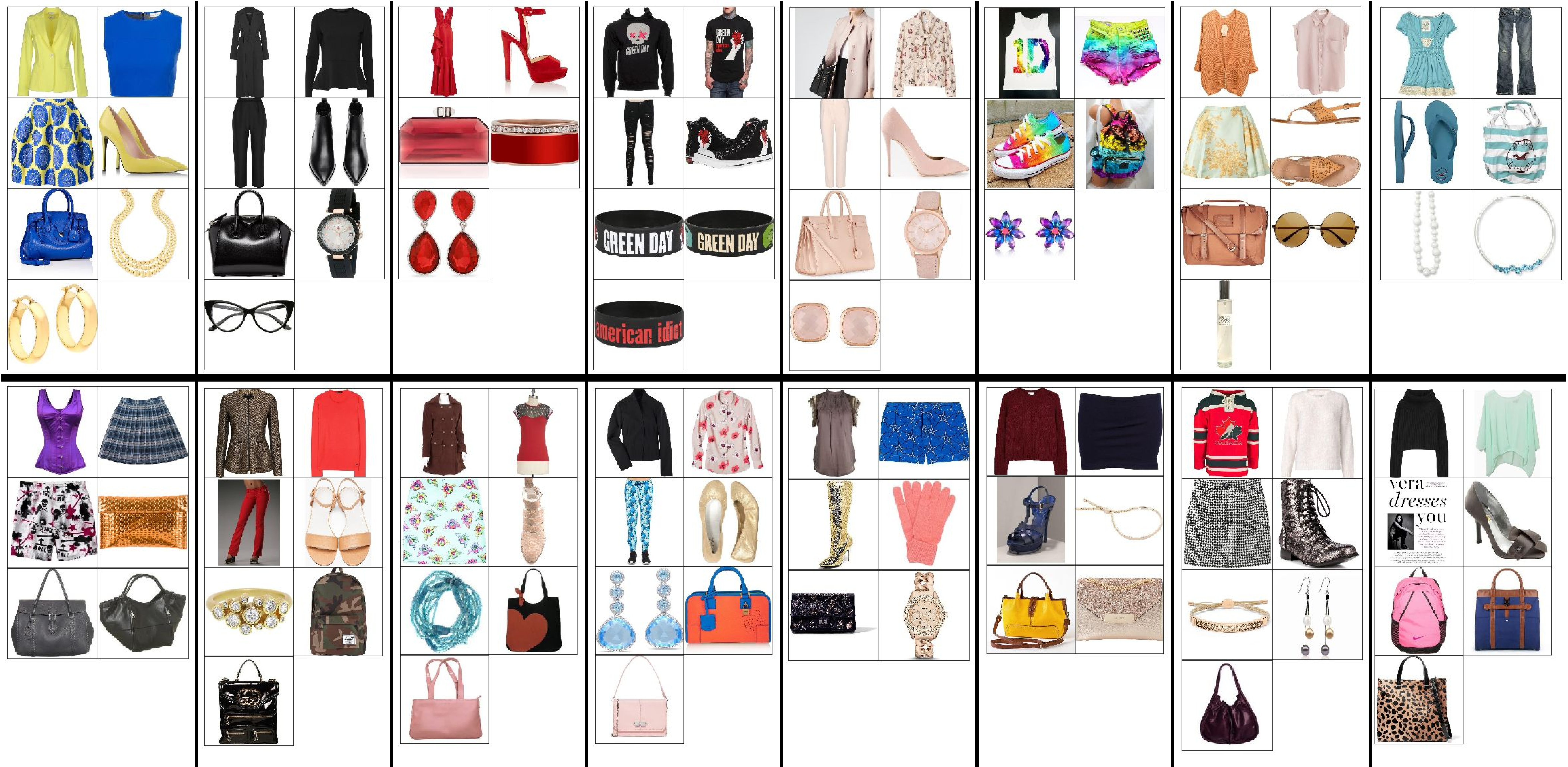}
  \caption{Eight best (top row) and worst (bottom row) outfits judged by our outfit grader}
  \label{fig:verifier_result}
  \vspace{-5pt}
\end{figure}

\subsection{Quantitative Results}
The accuracy, average precision, and average recall of all models are displayed in table~\ref{table:accuracy_precision}. According to the table, one\_fc4096 ,which has 84.51\% accuracy, 83.66\% average precision, and 80.62\% average recall, is clearly the best among the three models. The precision, recall and f1 value of both classes from one\_fc4096 model are shown in table~\ref{table:detail_metric}. Top 8 positive and negative samples from the model are shown in figure~\ref{fig:verifier_result}. Qualitatively, preferred outfits contain items with consistent colors and styles, whereas low-scoring outfits tend to have less common visual elements between items.

From table~\ref{table:detail_metric}, 92.29\% recall for negative class shows that the model is very reliable for pointing out the bad outfit. However, 68.95\% for the positive one shows that it tends to judge positive outfit as a negative one as well. When considering that the training performance is almost 100\% correct as shown in table~\ref{table:detail_metric}, we can conclude that the model overfits the training data.

\subsection{Color and Item Type Analysis}
We conduct another set of experiments to analyze the effect of various features on grading performance. We train one\_fc4096 for 100,000 iterations using 5 features: (1) item type, (2) 4-color palette, (3) (1)+(2), (4) ResNet-50 features from grayscale images, and (5) ResNet-50 features from RGB images. Item types are extracted from item name, and 4-color palettes are extracted from item image.

\renewcommand{\tabcolsep}{3pt}
\begin{table}
  \centering
  \small
\caption{Performances of one\_fc4096 outfit grader trained by different features: (1) item type, (2) 4-color palette, (3) (1)+(2), (4) ResNet-50 features from grayscale images, (5) ResNet-50 features from RGB images }
\label{table:feature_analysis}

\begin{tabular}{@{}lccc@{}}
\toprule
Feature& Accuracy & Avg. Precision & Avg. Recall \\ \midrule
(1) Item types & 74.33 & 71.77 & 67.02\\
(2) 4-color palettes & 74.53 & 72.71  & 66.35\\ 
(3) (1)+(2) & 78.93 & 77.57 & 73.03\\ 
(4) ResNet-50 grayscale & 81.31 & 79.35 & 77.69 \\ 
(5) ResNet-50 RGB & 84.26 & 83.06 & 80.73\\ \bottomrule
\end{tabular}
\end{table}

The result in table \ref{table:feature_analysis} shows that the item type and color represent the items equally, and the combination of them gives a better representation. However, the composite feature from ResNet-50 outperforms both primitive features, even without the color information. Finally, the color information in the ResNet-50 features affects the performance of outfit grader by 3\% classification accuracy.


\section{Human Evaluation} \label{sec:human_perception}

Outfit quality is a subjective topic. An outfit that looks chic to one person may look ugly to another. Although an evaluation on testing samples is important, we argue that it might be insufficient to verify the reliability of the approach. We conduct a large-scale human perception evaluation to further assess our model. We use the predictions from one\_fc4096 to do evaluations on human perception using Amazon Mechanical Turk (AMT).

\subsection{Geographic Trends}
People from different regions have different tastes in fashion. Since the one\_fc4096 model learned from data from \url{polyvore.com}, in order to show that the model successfully learned the compatibilities between fashion items, the model's predictions should be judged by people from the same region as the training data. After we inspect metadata of all 93,247 outfits that are used as positive samples, we found that they come from 39,590 different users. Around half of them (21,413 users ,54\%) did not provide the country. For the remaining 18,177 users, which come from 175 countries, most of them come from United States (8,167 users, 45\%), followed by Canada (872 users, 5\%), and other countries. As a result, our model's predictions will be judged by Americans.

\subsection{Evaluation Protocol} \label{sec:human_motivation}
We setup the experiment as choosing the better outfit from each pair to minimize the effect of absolute bias or personal preference from human subjects. In addition, outfits in each pair must have exactly same outfit parts at the same location in the outfit image, so that only the compatibility of items affects the judgments, not the outfits' configuration nor number of items in the outfits.

Our hypothesis is, 
if outfits in the pair have similar quality, people will choose both outfits equally. On the other hand, if the outfit has a large gap in quality, people will definitely choose one over the other. 
The quality score of each outfit come from our outfit grader. If our outfit grader can reliably judge the quality of the outfit, our hypothesis will be true.

To verify the hypothesis, we select a number of best outfits as references, denoted as \emph{Alpha ($\mathbf{A}$)}. Then, we select other outfits of different qualities, denoted as \emph{Delta ($\mathbf{\Delta}$)}, and pair them up with best outfits. After that, we show these pairs to human annotators. For each pair, we tell the annotators to choose the better of the two. 

Our expectation is that, the difference in quality between outfits in the $\mathbf{A}$ group and each of $\mathbf{\Delta}$ group is directly related to the probability that the annotators will choose outfits in the $\mathbf{A}$ group given the outfits in $\mathbf{\Delta}$. Since this experiment is set as pairwise comparisons, we believe that the mentioned probability should be approximated as
\begin{align}
p(s_\alpha|s_\delta) = s_\alpha/(s_\alpha+s_\delta)
\end{align}
where $s_\alpha$ and $s_\delta$ are positive probability calculated by our model of outfits in $\mathbf{A}$ and $\mathbf{\Delta}$, respectively.

\begin{figure}
\centering
  \includegraphics[width=.45\textwidth]{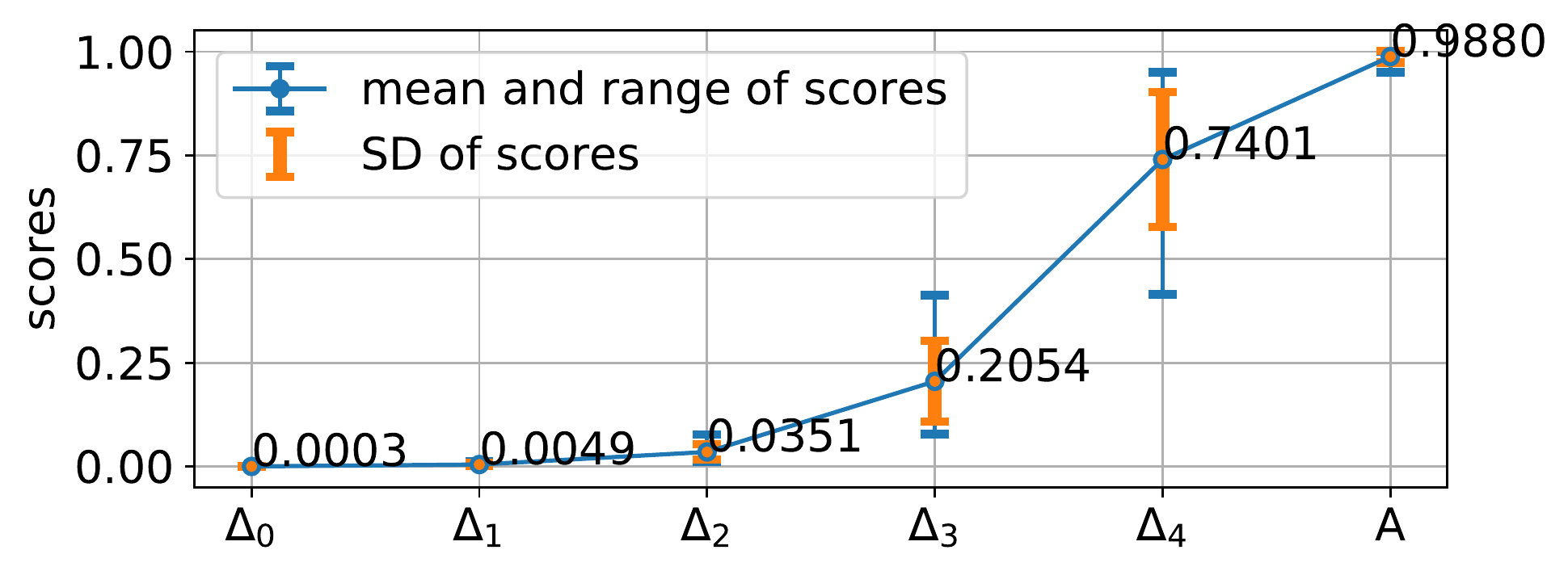}
   \caption{Mean, range, and SD of scores in each samples group $\mathbf{\Delta}$ and $\mathbf{A}$}
  \label{fig:mean_sd_score_deltas}
\end{figure}

\subsection{Implementation detail}
Our outfit's \emph{score} is the positive probability from the outfit grader.
We randomly select 1,000 outfits with
the score
more than 95\% as ``$\mathbf{A}$'' group. After that, we sort outfits with score less than 95\% in an ascending order, then divide them into 5 groups, from ``$\mathbf{\Delta_0}$'' which is the group of outfits with the lowest scores, to ``$\mathbf{\Delta_4}$'' which is outfit with the highest scores but still less than 95\%.

The experiment consists of 5,000 pairs of outfits. We use outfits from $\mathbf{A}$ group as ``good'' outfits, and $\mathbf{\Delta_j}$ for $j \in \{0,1,2,3,4\}$ as ``bad'' outfit. For each $\alpha_i \in \mathbf{A}$, we randomly select an outfit $\delta_{j,i} \in \mathbf{\Delta_j}$ for each $j \in \{0,1,2,3,4\}$ that has exactly same outfit parts as $\alpha_i$. Our hypothesis is, the visible difference in outfit quality in $(\alpha_i,\delta_{0,i})$ pairs is more than in $(\alpha_i,\delta_{4,i})$. We denote pair $(\alpha_i,\delta_{j,i})$ as $p_{j,i}$ for $j \in \{0,1,2,3,4\}$ and $i \in \{0,1,..,999\}$. We show in figure~\ref{fig:mean_sd_score_deltas} the mean, range, standard deviation of scores in each $\mathbf{\Delta_j}$ for $j \in \{0,1,2,3,4\}$ and $\mathbf{A}$, and the difference of mean of scores of each group to $\mathbf{A}$.

\begin{figure}
\centering
  \includegraphics[width=.4\textwidth]{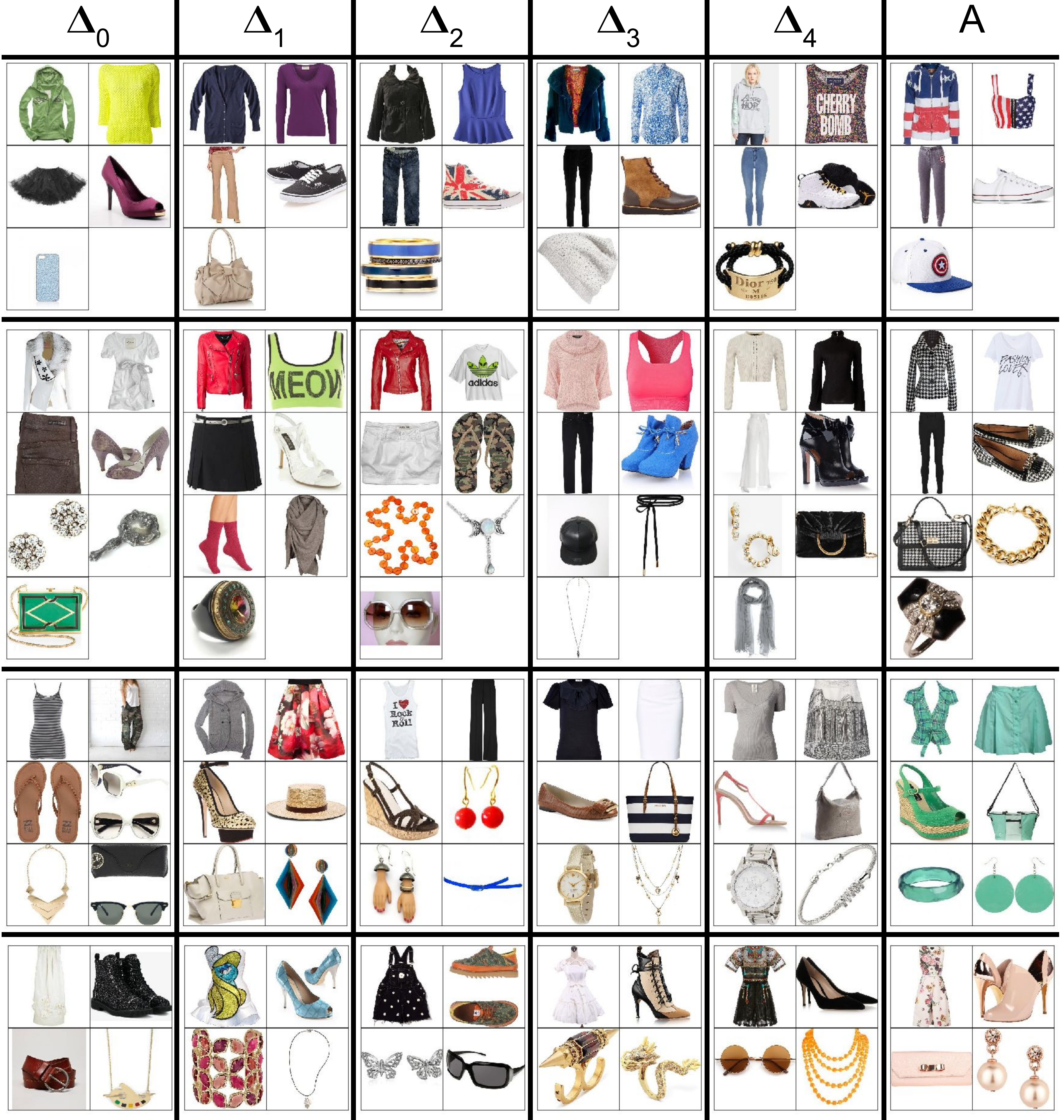}
  \caption{Comparison of outfits used in human evaluation. Each row shows outfits in different quality groups but have the same outfit configuration.}
  \label{fig:amt_outfits_examples}
  \vspace{-5pt}
\end{figure}

\begin{figure}
\centering
  \includegraphics[width=.4\textwidth]{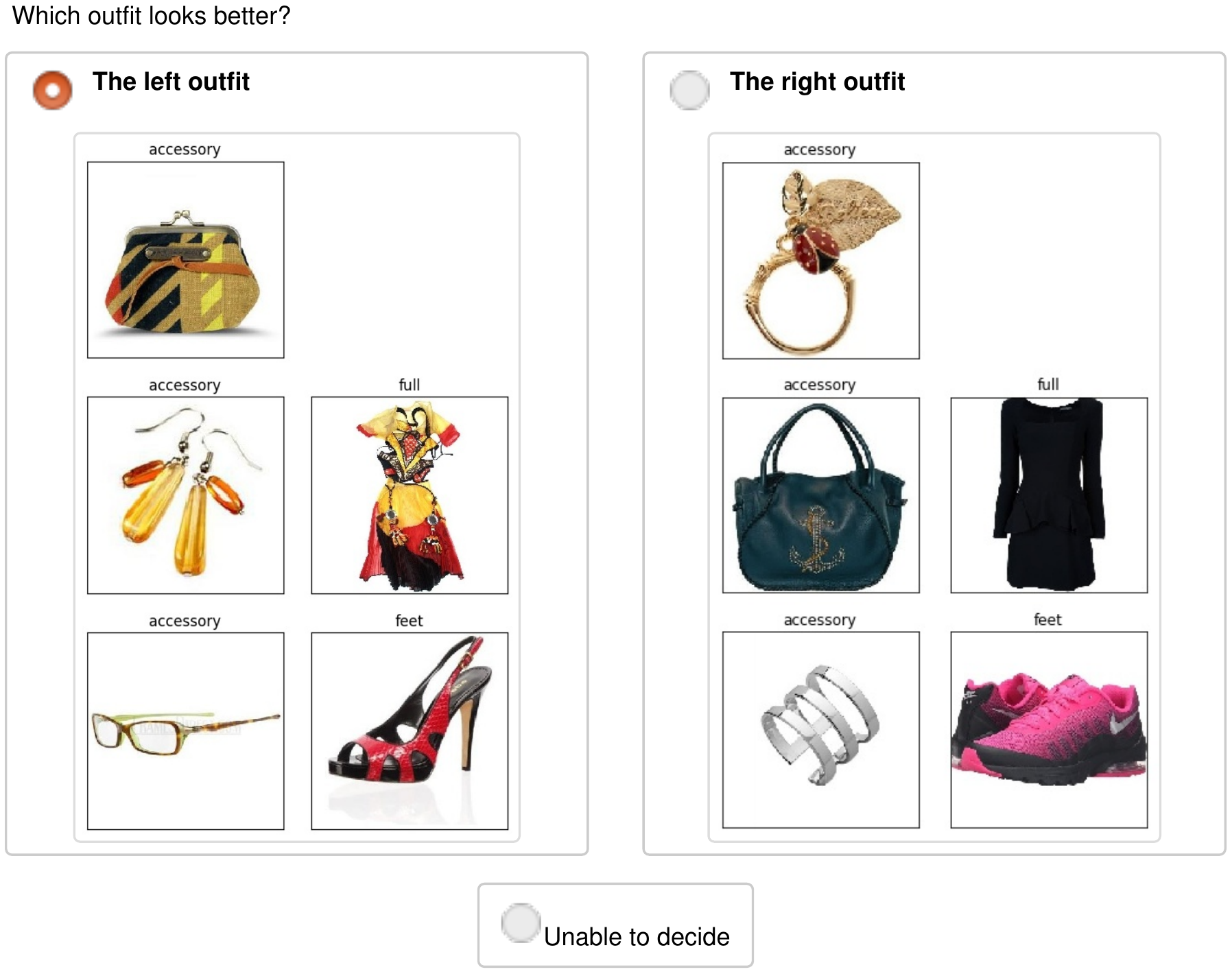}
  \vspace{-5pt}
  \caption{An example of questionnaires used in human evaluation, with associated outfit part of each item}
  \label{fig:questionnaire}
  \vspace{-5pt}
\end{figure}

We ask 5 annotators to vote each pair $p_{j,i}$. Each annotator selects the better outfit in each pair, or select ``Unable to decide'' if the annotator thinks that the outfits looks equally good (or bad). The total number of questions in our experiment equals to: 5 annotators $\times$ 1,000 questions $\times$ 5 $\delta$s = 25,000 questions. We show examples of outfits in figure~\ref{fig:amt_outfits_examples}. For each row, 5 pairs of outfits are created by pairing an outfit in $\mathbf{\Delta_j}$ for $j \in \{0,1,2,3,4\}$ with $\mathbf{A}$. An example questionnaire is shown in figure~\ref{fig:questionnaire}.

\begin{table}[t]
\small
\centering
\caption{Number of ``Unable to decide'' answers and ties from the experiments comparing outfits in $\mathbf{A}$ to $\mathbf{\Delta_j}$ for $j \in {0,1,2,3,4}$ }
\label{table:amt_result}

\begin{tabular}{@{}lccccc@{}}
\toprule
$j$& 0 & 1 & 2 & 3 & 4 \\ \midrule
\begin{tabular}[c]{@{}l@{}}Number of "Unable to decide"\\ (out of 5,000 questions)\end{tabular} & 688 & 820 & 924 & 696 & 422 \\ \midrule
\begin{tabular}[c]{@{}l@{}}Number of ties\\ (Out of 1,000 pairs)\end{tabular} & 63 & 80 & 114 & 111 & 80 \\ \bottomrule
\end{tabular}

\end{table}

\subsection{Evaluation Metrics}
We use the term \emph{Matching Ratio} to describe the ratio that human annotators select $\alpha_i$ in pair $(\alpha_i,\delta_{j,i})$ as the better-looking outfit. We also remove the ``Unable to decide'' votes, shown in table~\ref{table:amt_result}, from the calculation. There are two metrics, matching ratio by individual answer, and by majority vote on each pair. For the latter, we also remove ties, shown in table~\ref{table:amt_result}, from the calculation.

\subsection{Results}
\begin{figure}
\centering
  \includegraphics[width=.45\textwidth]{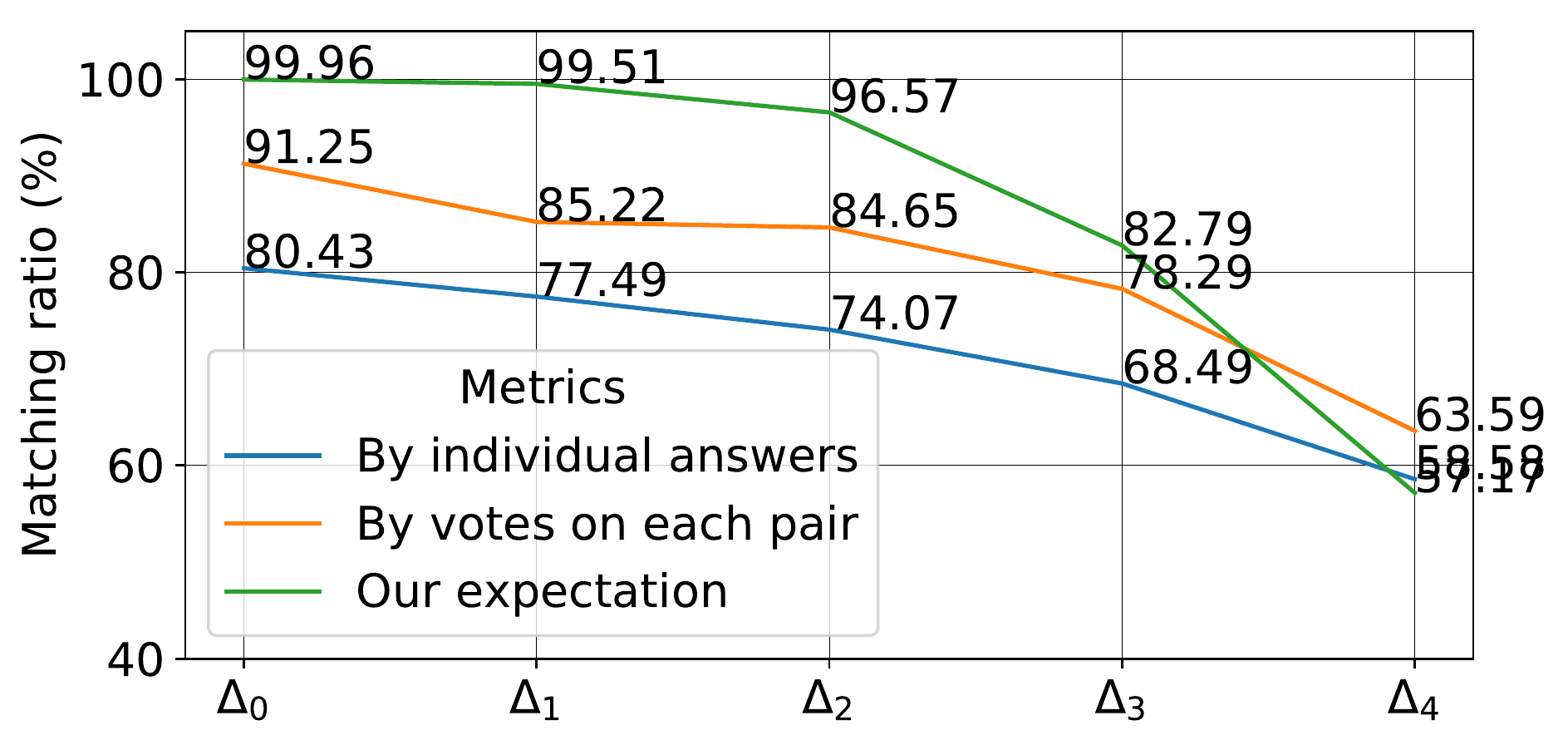}
  \caption{Matching ratio of the prediction to human judgment in each samples group $\mathbf{\Delta}$ with our expectation}
  \label{fig:amt_result}
  \vspace{-5pt}
\end{figure}

The results, with our expectation as explained in section~\ref{sec:human_motivation}, are shown in figure~\ref{fig:amt_result}. The 91.25\% matching ratio by voting shows that the human annotators agree with predictions from our model. Although not perfectly matched, the result has similar trend with our expectation. The result indicates that that the value of our positive probability (score) properly resembles the quality of the outfit.

Regarding the gap between human votes and our model, we have to remind that, the reliability of the human evaluations is not the absolute. As said earlier, fashion is a subjective topic. We might be able to use some small sets of questions to verify the ability of annotators, although this approach introduces absolute bias to the evaluation since people have different tastes in fashion.

\section{Application: Outfit Recommendation} \label{sec:application}




If the number of items is not very large, as is often the case with a personal closet, our outfit grader can directly be used as an outfit recommender by generating multiple outfits and ranking them. To demonstrate this usage, we conduct experiments as follows.

\subsection{Outfit generation}
For generating outfits, we consider four outfit configurations: (1) outer layer with upper- and lower-body, (2) only upper- and lower-body, (3) outer layer with full-body, and  (4) full-body only. All configurations include a footwear and at most three optional accessories.


Although it may be reasonable to assume that there are a modest number of clothes, there could be a large number of accessories and generating all possible item combinations is computationally expensive. We test four methods for generating outfits that take efficiency into consideration. The first method ({\em Ordered Beam Search}) is to regard  outfit generation  as a sequence generation problem, and employ a beam search (BS) algorithm. To be specific, in each item step $t$, the items that belongs to $part_t$ are the possible item extensions, and our ``one\_fc4096'' outfit grader is used as the scoring function. The BS starts from each item in the pool and considers all outfit configurations applicable to the item. It stops when all parts of the outfit are added according to its configuration. We then remove the duplicated outfits and recommend the best outfits based on the score from our outfit grader. 

The second method ({\em Orderless Beam Search}) uses the entire item pool as the possible item extensions at all time steps, while the rest is the same as the first one. The third method ({\em Partial Beam Search}) generates all possible combination of main parts (outer, upper, lower, full, feet) of the four outfit configurations, from which ten best outfits (based on score from our outfit grader) per outfit configuration per item are kept as base outfits. We then use the beam search to add accessories to those base outfits. The fourth method ({\em Baseline}) creates 100 outfits per outfit configuration in a random manner. Then, the duplicates are removed and the best ones are recommended.

Each of the four methods outputs 10 best outfits based on the scores from our outfit grader. In the experiments, for all the methods, we set the beam width for beam search to three and include a null item in the item pool as an accessory to give a choice to the beam search to add nothing to an outfit in each ``accessory'' time steps.






\subsection{Evaluation}
A good outfit recommender should be able to find sets of well-coordinate items in a pool of apparently random items. To test each recommendation method in terms of this property, we created 957 test cases, each of which contains items from one positive, denoted as $P$, and two negative samples. These samples are randomly drawn from the testing partition of Polyvore409k dataset. From those items, the recommended outfit, denoted as $R$, should be similar to the positive samples $P$. To measure the performance of the recommender, we use four conditions as 
(1) $P=R$, (2) $P\subset R$ (3) $R\subset P$, (4) $(P=R) \cup (P\subset R)\cup (R\subset P)$. For each method, we regard a recommendation (i.e., top ten recommended outfits) as successful if the condition is met by one of the ten recommended outfits.



\begin{figure}[t]
\centering
  \includegraphics[width=.45\textwidth]{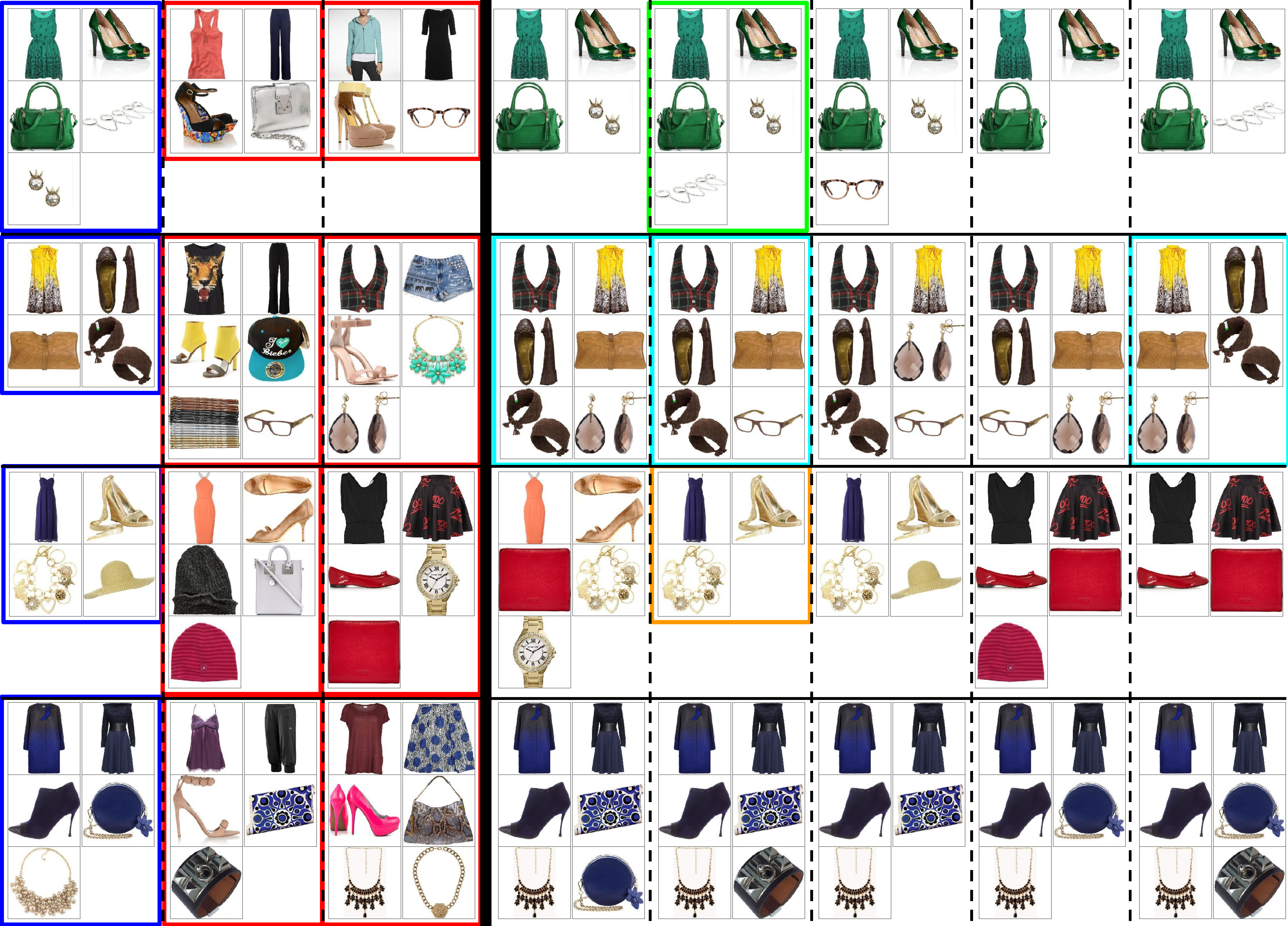}
  \caption{Recommended outfits from Partial BS. Each row shows one test case, where 5 outfits on the right are generated from items in 3 outfits on the left. Outfits with blue border are positive, red are negative, green are \emph{exact match}, cyan are $P\subset R$, and orange are $R\subset P$. The others are recommended outfits that do not meet the conditions.}
  \label{fig:recommended_outfits}
  \vspace{-5pt}
\end{figure}

Table~\ref{table:outfit_gen_eval} shows the results. It is seen that {\em Partial BS} outperforms the baseline in every metrics. The reason why Ordered and Orderless BS perform worse than Partial BS is because our outfit grader is trained using complete outfits, while the early steps of BS rely on the score of partial outfits, which our outfit grader is not trained for. Figure~\ref{fig:recommended_outfits} shows successful and unsuccessful recommendations. We argue that the recommended outfits in the failure case are even better than the target positive sample. This is due to the nature of weakly-supervised data.

\begin{table}[t]
\centering
\small
\caption{Performance of outfit recommendation by the proposed outfit grader combined with four outfit creation methods. The metrics are (1) $P=R$, (2) $P\subset R$ (3) $R\subset P$, (4) $(P=R) \cup (P\subset R)\cup (R\subset P)$, where $P$ and $R$ denote the positive sample and recommended outfits, respectively.}
\label{table:outfit_gen_eval}
\begin{tabular}{@{}lcccc@{}}
\toprule
Approaches & (1) & (2) & (3) & (4) \\ \midrule
Ordered BS & 11.39 & 14.11 & 19.64 & 32.29 \\
Orderless BS & 14.84 & 20.79 & 9.40 & 29.89 \\
\textbf{Partial BS} & \textbf{34.38} & \textbf{41.80} & \textbf{22.68} & \textbf{59.77} \\
Baseline & 8.88 & 21.53 & 14.11 & 36.36 \\ \bottomrule
\end{tabular}
\vspace{-5pt}
\end{table}

\section{Conclusion} \label{sec:conclusion}
In this paper, we study outfits as combinations of items by developing outfit graders and outfit recommenders. Given a combination of items as an outfit, our best model can judge if the outfit looks good or not at over 84\% accuracy on testing samples, and at 91\% matching ratio on evaluations by human annotators. In addition, user can just give a pool of items that user have to our outfit recommender, and it will recommend outfits from the item pool. We also collect a large clothing dataset consisting of over 600,000 clothing items and over 400,000 outfits, and use the dataset to learn and evaluate the outfit graders and recommenders.

\subsection*{Acknowledgements} \label{sec:acknowledgements}
This work was partly supported by JSPS KAKENHI Grant Number JP15H05919 and CREST, JST Grant Number JPMJCR14D1.

{\small
\bibliographystyle{ieeetr}
\bibliography{myrefs}
}

\end{document}